\documentclass[11pt,a4paper]{article}
\usepackage[hyperref]{conll-2019}
\usepackage{times}
\usepackage{latexsym}

\usepackage{url}

\usepackage{amsmath}
\usepackage{nccmath}
\usepackage{amssymb}
\usepackage{graphicx}
\usepackage{multirow}
\usepackage{subfig}

\aclfinalcopy 

\title{Effective Attention Modeling for Neural Relation Extraction}

\author{Tapas Nayak \and Hwee Tou Ng \\
 Department of Computer Science \\
 National University of Singapore \\
 nayakt@u.nus.edu, nght@comp.nus.edu.sg
}

\date{}

\begin{document}
\maketitle

\begin{abstract}
  Relation extraction is the task of determining the relation between two entities in a sentence. Distantly-supervised models are popular for this task. However, sentences can be long and two entities can be located far from each other in a sentence. The pieces of evidence supporting the presence of a relation between two entities may not be very direct, since the entities may be connected via some indirect links such as a third entity or via co-reference. Relation extraction in such scenarios becomes more challenging as we need to capture the long-distance interactions among the entities and other words in the sentence. Also, the words in a sentence do not contribute equally in identifying the relation between the two entities. To address this issue, we propose a novel and effective attention model which incorporates syntactic information of the sentence and a multi-factor attention mechanism. Experiments on the New York Times corpus show that our proposed model outperforms prior state-of-the-art models.
\end{abstract}

\section{Introduction}

Relation extraction from unstructured text is an important task to build knowledge bases (KB) automatically. \newcite{banko2007open} used open information extraction (Open IE) to extract relation triples from sentences where verbs were considered as the relation, whereas supervised information extraction systems extract a set of pre-defined relations from text. \newcite{mintz2009distant}, \newcite{riedel2010modeling}, and \newcite{hoffmann2011knowledge} proposed distant supervision to generate the training data for sentence-level relation extraction, where relation tuples (two entities and the relation between them) from a knowledge base such as Freebase \cite{bollacker2008freebase} were mapped to free text (Wikipedia articles or New York Times articles). The idea is that if a sentence contains both entities of a tuple, it is chosen as a training sentence of that tuple. Although this process can generate some noisy training instances, it can give a significant amount of training data which can be used to build supervised models for this task.

\newcite{mintz2009distant}, \newcite{riedel2010modeling}, and \newcite{hoffmann2011knowledge} proposed feature-based learning models and used entity tokens and their nearby tokens, their part-of-speech tags, and other linguistic features to train their models. Recently, many neural network-based models have been proposed to avoid feature engineering. \newcite{zeng2014relation,zeng2015distant} used convolutional neural networks (CNN) with max-pooling to find the relation between two given entities. Though these models have been shown to perform reasonably well on distantly supervised data, they sometimes fail to find the relation when sentences are long and entities are located far from each other. CNN models with max-pooling have limitations in understanding the semantic similarity of words with the given entities and they also fail to capture the long-distance dependencies among the words and entities such as co-reference. In addition, all the words in a sentence may not be equally important in finding the relation and this issue is more prominent in long sentences. Prior CNN-based models have limitations in identifying the multiple important factors to focus on in sentence-level relation extraction.

To address this issue, we propose a {\it novel} multi-factor attention model\footnote{The code and data of this paper can be found at https://github.com/nusnlp/MFA4RE} focusing on the syntactic structure of a sentence for relation extraction. We use a dependency parser to obtain the syntactic structure of a sentence. We use a linear form of attention to measure the semantic similarity of words with the given entities and combine it with the dependency distance of words from the given entities to measure their influence in identifying the relation. Also, single attention may not be able to capture all pieces of evidence for identifying the relation due to normalization of attention scores. Thus we use multi-factor attention in the proposed model. Experiments on the New York Times (NYT) corpus show that the proposed model outperforms prior work in terms of F1 scores on sentence-level relation extraction. 

\section{Task Description}

Sentence-level relation extraction is defined as follows: Given a sentence $S$ and two entities $\{E_1, E_2\}$ marked in the sentence, find the relation $r(E_1,E_2)$ between these two entities in $S$ from a pre-defined set of relations $R \cup \{\mathit{None}\}$. {\em None} indicates that none of the relations in $R$ holds between the two marked entities in the sentence. The relation between the entities is argument order-specific, i.e., $r(E_1,E_2)$ and $r(E_2,E_1)$ are not the same. Input to the system is a sentence $S$ and two entities $E_1$ and $E_2$, and output is the relation $r(E_1,E_2) \in R \cup \{\mathit{None}\}$.

\section{Model Description}

We use four types of embedding vectors in our model: (1) word embedding vector $\mathbf{w} \in \mathbb{R}^{d_w}$ (2) entity token indicator embedding vector $\mathbf{z} \in \mathbb{R}^{d_z}$, which indicates if a word belongs to entity $1$, entity $2$, or does not belong to any entity (3) a positional embedding vector $\mathbf{u}^1 \in \mathbb{R}^{d_u}$ which represents the linear distance of a word from the start token of entity $1$ (4) another positional embedding vector $\mathbf{u}^2 \in \mathbb{R}^{d_u}$ which represents the linear distance of a word from the start token of entity $2$.

We use a bi-directional long short-term memory (Bi-LSTM) \cite{hochreiter1997long} layer to capture the interaction among words in a sentence $S=\{w_1, w_2, ....., w_n\}$, where $n$ is the sentence length. The input to this layer is the concatenated vector $\mathbf{x} \in \mathbb{R}^{d_w + d_z}$ of word embedding vector $\mathbf{w}$ and entity token indicator embedding vector $\mathbf{z}$. 

\begin{align*}
&\mathbf{x}_t=\mathbf{w}_t ~||~ \mathbf{z}_t\\
&\overrightarrow{\mathbf{h}_t} = \overrightarrow{\mathrm{LSTM}}(\mathbf{x}_t, \mathbf{h}_{t-1}) \\
&\overleftarrow{\mathbf{h}_t} = \overleftarrow{\mathrm{LSTM}}(\mathbf{x}_t, \mathbf{h}_{t+1}) \\
&\mathbf{h}_t = \overrightarrow{\mathbf{h}_t} || \overleftarrow{\mathbf{h}_t}
\end{align*}

\noindent $\overrightarrow{\mathbf{h}_t} \in \mathbb{R}^{d_w+d_z}$ and $\overleftarrow{\mathbf{h}_t} \in \mathbb{R}^{d_w+d_z}$ are the output at the $t$th step of the forward LSTM and backward LSTM respectively. We concatenate them to obtain the $t$th Bi-LSTM output $\mathbf{h}_t \in \mathbb{R}^{2(d_w+d_z)}$.

\subsection{Global Feature Extraction}

We use a convolutional neural network (CNN) to extract the sentence-level global features for relation extraction. We concatenate the positional embeddings $\mathbf{u}^1$ and $\mathbf{u}^2$ of words with the hidden representation of the Bi-LSTM layer and use the convolution operation with max-pooling on concatenated vectors to extract the global feature vector. 
\begin{align*}
&\mathbf{q}_t = \mathbf{h}_t \Vert \mathbf{u}_t^1 \Vert \mathbf{u}_t^2\\
&c_i = \mathbf{f}^T (\mathbf{q}_{i} \Vert \mathbf{q}_{i + 1} \Vert .... \Vert \mathbf{q}_{i+k-1})\\
&c_{max} = \mathrm{max}(c_1,c_2,....,c_{n})\\
&\mathbf{v}_g = [c_{max}^1, c_{max}^2, ...., c_{max}^{f_g}]
\end{align*} 
$\mathbf{q}_t \in \mathbb{R}^{2(d_w+d_z+d_u)}$ is the concatenated vector for the $t$th word and $\mathbf{f}$ is a convolutional filter vector of dimension $2k(d_w+d_z+d_u)$ where $k$ is the filter width. The index $i$ moves from $1$ to $n$ and produces a set of scalar values $\{c_1, c_2, .....,c_{n}\}$. The max-pooling operation chooses the maximum $c_{max}$ from these values as a feature. With $f_g$ number of filters, we get a global feature vector $\mathbf{v}_g \in \mathbb{R}^{f_g}$.

\begin{figure*}[ht]
\centering
\includegraphics[scale=0.5]{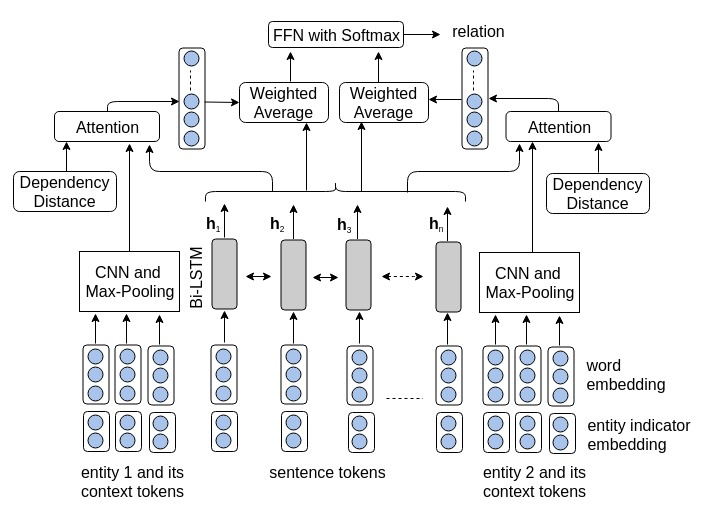}
\caption{Architecture of our attention model with $m=1$. We have not shown the CNN-based global feature extraction here. FFN=feed-forward network.}
\label{fig:joint-model}
\end{figure*}

\subsection{Attention Modeling}

Figure~\ref{fig:joint-model} shows the architecture of our attention model. We use a linear form of attention to find the semantically meaningful words in a sentence with respect to the entities which provide the pieces of evidence for the relation between them. Our attention mechanism uses the entities as attention queries and their vector representation is very important for our model. Named entities mostly consist of multiple tokens and many of them may not be present in the training data or their frequency may be low. The nearby words of an entity can give significant information about the entity. Thus we use the tokens of an entity and its nearby tokens to obtain its vector representation. We use the convolution operation with max-pooling in the context of an entity to get its vector representation.
\begin{align*}
&c_i = \mathbf{f}^T (\mathbf{x}_{i} \Vert \mathbf{x}_{i + 1} \Vert .... \Vert \mathbf{x}_{i+k-1})\\
&c_{max} = \mathrm{max}(c_b,c_{b+1},....,c_{e})\\
&\mathbf{v}_e = [c_{max}^1, c_{max}^2, ...., c_{max}^{f_e}]
\end{align*} 

\noindent $\mathbf{f}$ is a convolutional filter vector of size $k (d_w+d_z)$ where $k$ is the filter width and $\mathbf{x}$ is the concatenated vector of word embedding vector ($\mathbf{w}$) and entity token indicator embedding vector ($\mathbf{z}$). $b$ and $e$ are the start and end index of the sequence of words comprising an entity and its neighboring context in the sentence, where $1 \leq b \leq e \leq n$. The index $i$ moves from $b$ to $e$ and produces a set of scalar values $\{c_b, c_{b+1}, .....,c_{e}\}$. The max-pooling operation chooses the maximum $c_{max}$ from these values as a feature. With $f_e$ number of filters, we get the entity vector $\mathbf{v}_e \in \mathbb{R}^{f_e}$. We do this for both entities and get $\mathbf{v}_e^1 \in \mathbb{R}^{f_e}$ and $\mathbf{v}_e^2 \in \mathbb{R}^{f_e}$ as their vector representation. We adopt a simple linear function as follows to measure the semantic similarity of words with the given entities:
\begin{align*}
&\mathrm{f_{score}^1}(\mathbf{h}_i, \mathbf{v}_e^1)=\mathbf{h}_i^T \mathbf{W}_a^1 \mathbf{v}_e^1\\
&\mathrm{f_{score}^2}(\mathbf{h}_i, \mathbf{v}_e^2)=\mathbf{h}_i^T \mathbf{W}_a^2 \mathbf{v}_e^2
\end{align*}
$\mathbf{h}_i$ is the Bi-LSTM hidden representation of the $i$th word. $\mathbf{W}_a^1, \mathbf{W}_a^2 \in \mathbb{R}^{2(d_w+d_z) \times f_e}$ are trainable weight matrices. $\mathrm{f_{score}^1}(\mathbf{h}_i, \mathbf{v}_e^1)$ and $\mathrm{f_{score}^2}(\mathbf{h}_i, \mathbf{v}_e^2)$ represent the semantic similarity score of the $i$th word and the two given entities.

\begin{figure}[t]
\centering
\includegraphics[scale=0.4]{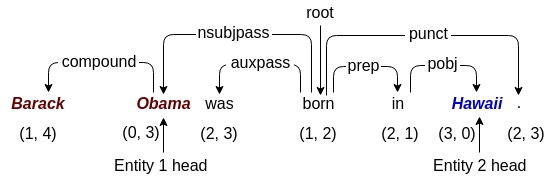}
\caption{An example dependency tree. The two numbers indicate the distance of the word from the head token of the two entities respectively along the dependency tree path.}
\label{fig:dep_dist}
\end{figure}

Not all words in a sentence are equally important in finding the relation between the two entities. The words which are closer to the entities are generally more important. To address this issue, we propose to incorporate the syntactic structure of a sentence in our attention mechanism. The syntactic structure is obtained from the dependency parse tree of the sentence. Words which are closer to the entities in the dependency parse tree are more relevant to finding the relation. In our model, we define the dependency distance to every word from the head token  (last token) of an entity as the number of edges along the dependency path (See Figure~\ref{fig:dep_dist} for an example). We use a distance window size $ws$ and words whose dependency distance is within this window receive attention and the other words are ignored. The details of our attention mechanism follow.
\begin{align*}
&d_i^1 = \begin{cases}
    \frac{1}{2^{l_i^1-1}} \mathrm{exp(f_{score}^1}(\mathbf{h}_i, \mathbf{v}_e^1)) & \text{if } l_i^1 \in [1,ws]\\
    \frac{1}{2^{ws}} \mathrm{exp(f_{score}^1}(\mathbf{h}_i, \mathbf{v}_e^1)) & \text{otherwise}
\end{cases} \\
&d_i^2 = \begin{cases}
    \frac{1}{2^{l_i^2-1}} \mathrm{exp(f_{score}^2}(\mathbf{h}_i, \mathbf{v}_e^2)) & \text{if } l_i^2 \in [1,ws]\\
    \frac{1}{2^{ws}} \mathrm{exp(f_{score}^2}(\mathbf{h}_i, \mathbf{v}_e^2)) & \text{otherwise}
\end{cases}\\
&p_i^1 = \frac{d_i^1}{\sum_j{d_j^1}}, \quad \quad \quad \quad p_i^2 = \frac{d_i^2}{\sum_j{d_j^2}}
\end{align*}
$d_i^1$ and $d_i^2$ are un-normalized attention scores and $p_i^1$ and $p_i^2$ are the normalized attention scores for the $i$th word with respect to entity 1 and entity 2 respectively. $l_i^1$ and $l_i^2$ are the dependency distances of the $i$th word from the two entities. We mask those words whose average dependency distance from the two entities is larger than $ws$. We use the semantic meaning of the words and their dependency distance from the two entities together in our attention mechanism. The attention feature vectors $\mathbf{v}_a^1$ and $\mathbf{v}_a^1$ with respect to the two entities are determined as follows:
\begin{align*}
    &\mathbf{v}_a^1 = \sum_{i=1}^n p_i^1 \mathbf{h}_i, \quad \quad \mathbf{v}_a^2 = \sum_{i=1}^n p_i^2 \mathbf{h}_i 
\end{align*}

\subsection{Multi-Factor Attention}

Two entities in a sentence, when located far from each other, can be linked via more than one co-reference chain or more than one important word. Due to the normalization of the attention scores as described above, single attention cannot capture all relevant information needed to find the relation between two entities. Thus we use a multi-factor attention mechanism, where the number of factors is a hyper-parameter, to gather all relevant information for identifying the relation. We replace the attention matrix $\mathbf{W}_a$ with an attention tensor $\mathbf{W}_a^{1:m} \in \mathbb{R}^{m \times 2(d_w+d_z) \times 2f_e}$ where $m$ is the factor count. This gives us $m$ attention vectors with respect to each entity. We concatenate all the feature vectors obtained using these attention vectors to get the multi-attentive feature vector $\mathbf{v}_{ma} \in \mathbb{R}^{4 m (d_w+d_z)}$. 

\subsection{Relation Extraction}

We concatenate $\mathbf{v}_g$, $\mathbf{v}_{ma}$, $\mathbf{v}_e^1$, and $\mathbf{v}_e^2$, and this concatenated feature vector is given to a feed-forward layer with softmax activation to predict the normalized probabilities for the relation labels.
\begin{align*}
&\mathbf{r} = \mathrm{softmax}(\mathbf{W}_r (\mathbf{v}_g~||~\mathbf{v}_{ma} ~||~ \mathbf{v}_e^1 ~||~ \mathbf{v}_e^2)+\mathbf{b}_r)
\end{align*}
$\mathbf{W}_r \in \mathbb{R}^{(f_g + 2 f_e + 4 m (d_w+d_z)) \times (\vert R \vert+1)}$ is the weight matrix, $\mathbf{b}_r \in \mathbb{R}^{\vert R \vert+1}$ is the bias vector of the feed-forward layer for relation extraction, and $\mathbf{r}$ is the vector of normalized probabilities of relation labels.

\subsection{Loss Function}
We calculate the loss over each mini-batch of size $B$. We use the following negative log-likelihood as our objective function for relation extraction:
\begin{equation*}
\begin{split}
\mathcal{L} = -\frac{1}{B} \sum_{i=1}^{B} \mathrm{log} (p(r_{i} \vert s_i, e_i^1, e_i^2, \theta))
\end{split}
\end{equation*}
where $p(r_{i} \vert s_i, e_i^1, e_i^2, \theta)$ is the conditional probability of the true relation $r_i$ when the sentence $s_i$, two entities $e_i^1$ and $e_i^2$, and the model parameters $\theta$ are given.

\section{Experiments}

\subsection{Datasets}

We use the New York Times (NYT) corpus \cite{riedel2010modeling} in our experiments. There are two versions of this corpus: (1) The original NYT corpus created by \newcite{riedel2010modeling} which has $52$ valid relations and a {\em None} relation. We name this dataset NYT10. The training dataset has $455,412$ instances and $330,776$ of the instances belong to the {\em None} relation and the remaining $124,636$ instances have valid relations. The test dataset has $172,415$ instances and $165,974$ of the instances belong to the {\em None} relation and the remaining $6,441$ instances have valid relations. Both the training and test datasets have been created by aligning Freebase \cite{bollacker2008freebase} tuples to New York Times articles. (2) Another version created by \newcite{hoffmann2011knowledge} which has $24$ valid relations and a {\em None} relation. We name this dataset NYT11. The corresponding statistics for NYT11 are given in Table~\ref{tab:dataset}. The training dataset is created by aligning Freebase tuples to NYT articles, but the test dataset is manually annotated.

\begin{table}[ht]
\small
\centering
\begin{tabular}{l|lcc}
\hline
 &  & \multicolumn{1}{l}{NYT10} & \multicolumn{1}{l}{NYT11} \\ \hline
 & \# relations & 53 & 25 \\ \hline
\multirow{5}{*}{Train} & \# instances & 455,412 & 335,843 \\  
 & \# valid relation tuples & 124,636 & 100,671 \\ 
 & \# None relation tuples & 330,776 & 235,172 \\  
 & avg. sentence length & 41.1 & 37.2 \\  
 & \begin{tabular}[c]{@{}l@{}}avg. distance between \\ entity pairs\end{tabular} & 12.8 & 12.2 \\ \hline
\multirow{5}{*}{Test} & \# instances & 172,415 & 1,450 \\  
 & \# valid relation tuples & 6,441 & 520 \\ 
 & \# None relation tuples & 165,974 & 930 \\  
 & avg. sentence length & 41.7 & 39.7 \\  
 & \begin{tabular}[c]{@{}l@{}}avg. distance between\\  entity pairs\end{tabular} & 13.1 & 11.0 \\ \hline
\end{tabular}
\caption{Statistics of the NYT10 and NYT11 dataset.}
\label{tab:dataset}
\end{table}

\subsection{Evaluation Metrics}

We use precision, recall, and F1 scores to evaluate the performance of models on relation extraction after removing the {\em None} labels. We use a confidence threshold to decide if the relation of a test instance belongs to the set of relations $R$ or {\em None}. If the network predicts {\em None} for a test instance, then it is considered as {\em None} only. But if the network predicts a relation from the set $R$ and the corresponding softmax score is below the confidence threshold, then the final class is changed to {\em None}. This confidence threshold is the one that achieves the highest F1 score on the validation dataset. We also include the precision-recall curves for all the models.

\subsection{Parameter Settings}

We run word2vec \cite{mikolov2013efficient} on the NYT corpus to obtain the initial word embeddings with dimension $d_w=50$ and update the embeddings during training. We set the dimension of entity token indicator embedding vector $d_z=10$ and positional embedding vector $d_u=5$. The hidden layer dimension of the forward and backward LSTM is $60$, which is the same as the dimension of input word representation vector $\mathbf{x}$. The dimension of Bi-LSTM output is $120$. We use $f_g=f_e=230$ filters of width $k=3$ for feature extraction whenever we apply the convolution operation. We use dropout in our network with a dropout rate of $0.5$, and in convolutional layers, we use the tanh activation function. We use the sequence of tokens starting from $5$ words before the entity to $5$ words after the entity as its context. We train our models with mini-batch size of $50$ and optimize the network parameters using the Adagrad optimizer \cite{duchi2011adaptive}. We use the dependency parser from spaCy\footnote{https://spacy.io/} to obtain the dependency distance of the words from the entities and use $ws=5$ as the window size for dependency distance-based attention.

\begin{table*}[t]
\centering
\begin{tabular}{llll|lll}
\hline \hline
 & \multicolumn{3}{c|}{NYT10} & \multicolumn{3}{c}{NYT11} \\ 
Model & Prec. & Rec. & F1 & \multicolumn{1}{c}{Prec.} & \multicolumn{1}{c}{Rec.} & \multicolumn{1}{c}{F1} \\ \hline
CNN \cite{zeng2014relation}  & 0.413 & 0.591 & 0.486 & 0.444 & 0.625 & 0.519  \\
PCNN \cite{zeng2015distant}  & 0.380 & \textbf{0.642} & 0.477 & 0.446 & 0.679 & 0.538$^\dagger$  \\
EA \cite{huang2016attention}  & 0.443 & 0.638 & 0.523$^\dagger$ & 0.419 & 0.677 & 0.517  \\ 
BGWA \cite{jat2018attention} & 0.364 & 0.632 & 0.462 & 0.417 & \textbf{0.692} & 0.521  \\ \hline
BiLSTM-CNN & 0.490 & 0.507 & 0.498 & 0.473 & 0.606 & 0.531  \\
Our model & \textbf{0.541} & 0.595 & \textbf{0.566}* & \textbf{0.507} & 0.652 & \textbf{0.571}*  \\ \hline \hline
\end{tabular}
\caption{Performance comparison of different models on the two datasets. * denotes a statistically significant improvement over the previous best state-of-the-art model with $p < 0.01$ under the bootstrap paired t-test. $^\dagger$ denotes the previous best state-of-the-art model.}
\label{tab:re}
\end{table*}

\subsection{Comparison to Prior Work}

We compare our proposed model with the following state-of-the-art models.

(1) CNN \cite{zeng2014relation}: Words are represented using word embeddings and two positional embeddings. A convolutional neural network (CNN) with max-pooling is applied to extract the sentence-level feature vector. This feature vector is passed to a feed-forward layer with softmax to classify the relation.

(2) PCNN \cite{zeng2015distant}: Words are represented using word embeddings and two positional embeddings. A convolutional neural network (CNN) is applied to the word representations. Rather than applying a global max-pooling operation on the entire sentence, three max-pooling operations are applied on three segments of the sentence based on the location of the two entities (hence the name Piecewise Convolutional Neural Network (PCNN)). The first max-pooling operation is applied from the beginning of the sentence to the end of the entity appearing first in the sentence. The second max-pooling operation is applied from the beginning of the entity appearing first in the sentence to the end of the entity appearing second in the sentence. The third max-pooling operation is applied from the beginning of the entity appearing second in the sentence to the end of the sentence. Max-pooled features are concatenated and passed to a feed-forward layer with softmax to determine the relation. 

(3) Entity Attention (EA) \cite{huang2016attention}: This is the combination of a CNN model and an attention model. Words are represented using word embeddings and two positional embeddings. A CNN with max-pooling is used to extract global features. Attention is applied with respect to the two entities separately. The vector representation of every word is concatenated with the word embedding of the last token of the entity. This concatenated representation is passed to a feed-forward layer with tanh activation and then another feed-forward layer to get a scalar attention score for every word. The original word representations are averaged based on the attention scores to get the attentive feature vectors. A CNN-extracted feature vector and two attentive feature vectors with respect to the two entities are concatenated and passed to a feed-forward layer with softmax to determine the relation.

(4) BiGRU Word Attention (BGWA) \cite{jat2018attention}: Words are represented using word embeddings and two positional embeddings. They are passed to a bidirectional gated recurrent unit (BiGRU) \cite{cho2014properties} layer. Hidden vectors of the BiGRU layer are passed to a bilinear operator (a combination of two feed-forward layers) to compute a scalar attention score for each word. Hidden vectors of the BiGRU layer are multiplied by their corresponding attention scores. A piece-wise CNN is applied on the weighted hidden vectors to obtain the feature vector. This feature vector is passed to a feed-forward layer with softmax to determine the relation.

(5) BiLSTM-CNN: This is our own baseline. Words are represented using word embeddings and entity indicator embeddings. They are passed to a bidirectional LSTM. Hidden representations of the LSTMs are concatenated with two positional embeddings. We use CNN and max-pooling on the concatenated representations to extract the feature vector. Also, we use CNN and max-pooling on the word embeddings and entity indicator embeddings of the context words of entities to obtain entity-specific features. These features are concatenated and passed to a feed-forward layer to determine the relation. 

\begin{figure}[t]
\centering
\includegraphics[scale=0.6]{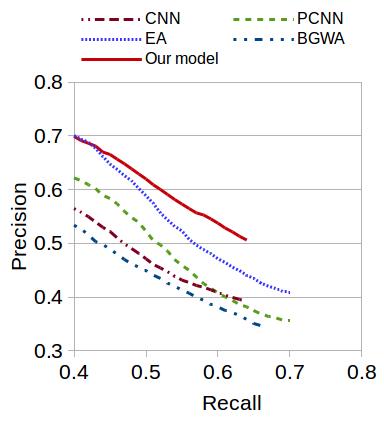}
\caption{Precision-Recall curve for the NYT10 dataset.}
\label{fig:nyt10_pr}
\end{figure}

\begin{figure}[t]
\centering
\includegraphics[scale=0.6]{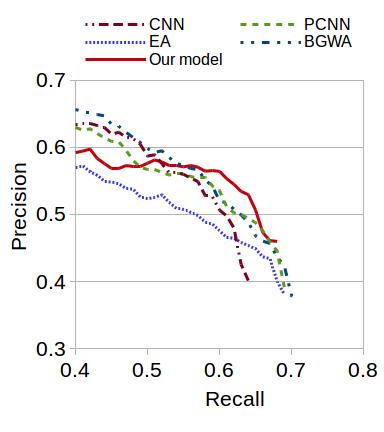}
\caption{Precision-Recall curve for the NYT11 dataset.}
\label{fig:nyt11_pr}
\end{figure}

\subsection{Experimental Results}

We present the results of our final model on the relation extraction task on the two datasets in Table \ref{tab:re}. Our model outperforms the previous state-of-the-art models on both datasets in terms of F1 score. On the NYT10 dataset, it achieves $4.3\%$ higher F1 score compared to the previous best state-of-the-art model EA. Similarly, it achieves $3.3\%$ higher F1 score compared to the previous best state-of-the-model PCNN on the NYT11 dataset. Our model improves the precision scores on both datasets with good recall scores. This will help to build a cleaner knowledge base with fewer false positives. We also show the precision-recall curves for the NYT10 and NYT11 datasets in Figures \ref{fig:nyt10_pr} and \ref{fig:nyt11_pr} respectively. The goal of any relation extraction system is to extract as many relations as possible with minimal false positives. If the recall score becomes very low, the coverage of the KB will be poor. From Figure \ref{fig:nyt10_pr}, we observe that when the recall score is above $0.4$, our model achieves higher precision than all the competing models on the NYT10 dataset. On the NYT11 dataset (Figure \ref{fig:nyt11_pr}), when recall score is above $0.6$, our model achieves higher precision than the competing models. Achieving higher precision with high recall score helps to build a cleaner KB with good coverage.

\begin{figure}[t]
\centering
\includegraphics[scale=0.6]{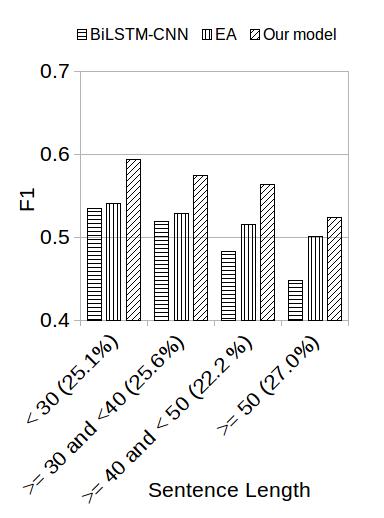}
\caption{Performance comparison across different sentence lengths on the NYT10 dataset.}
\label{fig:nyt10_len}
\end{figure}

\begin{figure}[t]
\centering
\includegraphics[scale=0.6]{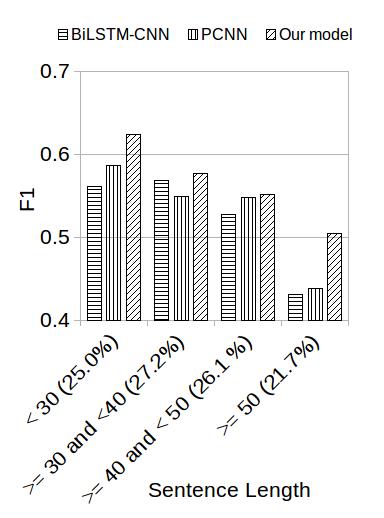}
\caption{Performance comparison across different sentence lengths on the NYT11 dataset.}
\label{fig:nyt11_len}
\end{figure}

\section{Analysis and Discussion}
\subsection{Varying the number of factors ($m$)}

We investigate the effects of the multi-factor count $(m)$ in our final model on the test datasets in Table \ref{tab:mfa}. We observe that for the NYT10 dataset, $m=\{1, 2, 3\}$ gives good performance with $m=1$ achieving the highest F1 score. On the NYT11 dataset, $m=4$ gives the best performance. These experiments show that the number of factors giving the best performance may vary depending on the underlying data distribution.

\begin{table}[ht]
\small
\centering
\begin{tabular}{llll|lll}
\hline \hline
 & \multicolumn{3}{c|}{NYT10} & \multicolumn{3}{c}{NYT11} \\ 
$m$ & Prec. & Rec. & F1 & \multicolumn{1}{c}{Prec.} & \multicolumn{1}{c}{Rec.} & \multicolumn{1}{c}{F1} \\ \hline
${1}$ & 0.541 & 0.595 & \textbf{0.566} & 0.495 & 0.621 & 0.551  \\
${2}$ & 0.521 & 0.597 & 0.556 & 0.482 & 0.656 & 0.555  \\
${3}$ & 0.490 & 0.617 & 0.547 & 0.509 & 0.633 & 0.564  \\ 
${4}$ & 0.449 & 0.623 & 0.522 & 0.507 & 0.652 & \textbf{0.571}  \\
${5}$ & 0.467 & 0.609 & 0.529 & 0.488 & 0.677 & 0.567  \\\hline \hline
\end{tabular}
\caption{Performance comparison of our model with different values of $m$ on the two datasets.}
\label{tab:mfa}
\end{table}

\begin{figure}[t]
\centering
\includegraphics[scale=0.57]{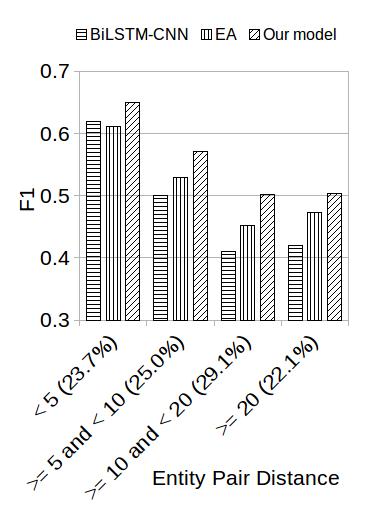}
\caption{Performance comparison across different distances between entities on the NYT10 dataset.}
\label{fig:nyt10_dist}
\end{figure}

\begin{figure}[t]
\centering
\includegraphics[scale=0.57]{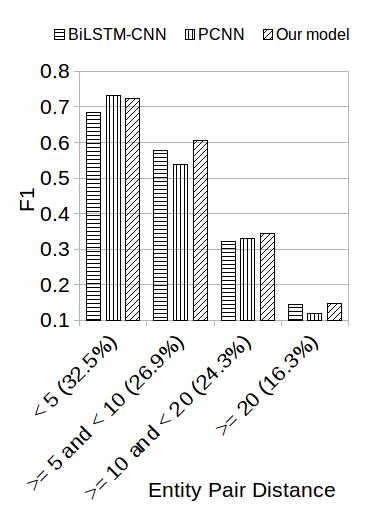}
\caption{Performance comparison across different distances between entities on the NYT11 dataset.}
\label{fig:nyt11_dist}
\end{figure}

\subsection{Effectiveness of Model Components}

We include the ablation results on the NYT11 dataset in Table \ref{tab:nyt11_ablation}. When we add multi-factor attention to the baseline BiLSTM-CNN model without the dependency distance-based weight factor in the attention mechanism, we get $0.8\%$ F1 score improvement (A2$-$A1). Adding the dependency weight factor with a window size of $5$ improves the F1 score by $3.2\%$ (A3$-$A2). Increasing the window size to $10$ reduces the F1 score marginally (A3$-$A4). Replacing the attention normalizing function with softmax operation also reduces the F1 score marginally (A3$-$A5). In our model, we concatenate the features extracted by each attention layer. Rather than concatenating them, we can apply max-pooling operation across the multiple attention scores to compute the final attention scores. These max-pooled attention scores are used to obtain the weighted average vector of Bi-LSTM hidden vectors. This affects the model performance negatively and F1 score of the model decreases by $3.0\%$ (A3$-$A6).

\begin{table}[ht]
\small
\centering
\begin{tabular}{lccc}
\hline
 & Prec. & Rec. & F1 \\ \hline
(A1) BiLSTM-CNN & 0.473 & 0.606 & 0.531 \\
(A2) Standard attention & 0.466 & 0.638 & 0.539 \\
(A3) Window size $(ws)=5$ & 0.507 & 0.652 & \textbf{0.571} \\
(A4) Window size $(ws)=10$ & 0.510 & 0.640 & 0.568 \\ 
(A5) Softmax & 0.490 & 0.658 & 0.562 \\ 
(A6) Max-pool & 0.492 & 0.600 & 0.541 \\ \hline
\end{tabular}
\caption{Effectiveness of model components ($m=4$) on the NYT11 dataset.}
\label{tab:nyt11_ablation}
\end{table}

\subsection{Performance with Varying Sentence Length and Varying Entity Pair Distance}

We analyze the effects of our attention model with different sentence lengths in the two datasets in Figures \ref{fig:nyt10_len} and \ref{fig:nyt11_len}. We also analyze the effects of our attention model with different distances between the two entities in the two datasets in Figures \ref{fig:nyt10_dist} and \ref{fig:nyt11_dist}. We observe that with increasing sentence length and increasing distance between the two entities, the performance of all models drops. This shows that finding the relation between entities located far from each other is a more difficult task. Our multi-factor attention model with dependency-distance weight factor increases the F1 score in all configurations when compared to previous state-of-the-art models on both datasets.

\section{Related Work} 

Relation extraction from a distantly supervised dataset is an important task and many researchers \cite{mintz2009distant, riedel2010modeling, hoffmann2011knowledge} tried to solve this task using feature-based classification models. Recently, \newcite{zeng2014relation, zeng2015distant} used CNN models for this task which can extract features automatically. \newcite{huang2016attention} and \newcite{jat2018attention} used attention mechanism in their model to improve performance. \newcite{mimlre}, \newcite{lin2016neural}, \newcite{vashishth2018reside}, \newcite{wu2018improving}, and \newcite{ye2019intra} used multiple sentences in a multi-instance relation extraction setting to capture the features located in multiple sentences for a pair of entities. In their evaluation setting, they evaluated model performance by considering multiple sentences having the same pair of entities as a single test instance. On the other hand, our model and the previous models that we compare to in this paper \cite{zeng2014relation,zeng2015distant,huang2016attention,jat2018attention} work on each sentence independently and are evaluated at the sentence level. Since there may not be multiple sentences that contain a pair of entities, it is important to improve the task performance at the sentence level. Future work can explore the integration of our sentence-level attention model in a multi-instance relation extraction framework. 

Not much previous research has exploited the dependency structure of a sentence in different ways for relation extraction. \newcite{Xu2015ClassifyingRV} and \newcite{miwa2016end} used an LSTM network and the shortest dependency path between two entities to find the relation between them. \newcite{Huang2017ImprovingSF} used the dependency structure of a sentence for the slot-filling task which is close to the relation extraction task. \newcite{liu-etal-2015-dependency} exploited the shortest dependency path between two entities and the sub-trees attached to that path (augmented dependency path) for relation extraction. \newcite{zhang-etal-2018-graph} and \newcite{guo2019aggcn} used graph convolution networks with pruned dependency tree structures for this task. In this work, we have incorporated the dependency distance of the words in a sentence from the two entities in a multi-factor attention mechanism to improve sentence-level relation extraction.

Attention-based neural networks are quite successful for many other NLP tasks. \newcite{Bahdanau2014NeuralMT} and \newcite{luong2015effective} used attention models for neural machine translation, \newcite{seo2016bidirectional} used attention mechanism for answer span extraction. \newcite{vaswani2017attention} and \newcite{kundu2018amanda} used multi-head or multi-factor attention models for machine translation and answer span extraction respectively. \newcite{effective2017ruidan} used dependency distance-focused word attention model for aspect-based sentiment analysis. 

\section{Conclusion}

In this paper, we have proposed a multi-factor attention model utilizing syntactic structure for relation extraction. The syntactic structure component of our model helps to identify important words in a sentence and the multi-factor component helps to gather different pieces of evidence present in a sentence. Together, these two components improve the performance of our model on this task, and our model outperforms previous state-of-the-art models when evaluated on the New York Times (NYT) corpus, achieving significantly higher F1 scores.

\section*{Acknowledgments}

We would like to thank the anonymous reviewers for their valuable and constructive comments on this work.

\bibliography{conll-2019}
\bibliographystyle{acl_natbib}

\end{document}